\documentclass[11pt]{article}
\usepackage{coling2020}
\usepackage{times}
\usepackage{url}
\usepackage{latexsym}

\usepackage{multirow}
\usepackage{booktabs}
\usepackage{graphicx}
\graphicspath{{figure/}}

\usepackage[american]{babel}
\usepackage{epstopdf}
\usepackage{multirow}
\usepackage{float}
\usepackage{bm}
\usepackage{amssymb}
\usepackage{amsmath}
\usepackage{subfigure}
\usepackage{xcolor}
\usepackage[linesnumbered,ruled,lined,commentsnumbered]{algorithm2e}

\colingfinalcopy

\title{Early Detection of Fake News by Utilizing the Credibility of News, Publishers, and Users Based on Weakly Supervised Learning}

\author{
	Chunyuan Yuan\textsuperscript{\rm 1,2}, Qianwen Ma\textsuperscript{\rm 1,2}, Wei Zhou\textsuperscript{\rm 1,*}, \textbf{Jizhong Han\textsuperscript{\rm 2} and Songlin Hu\textsuperscript{\rm 1,2,*} } \\
	\textsuperscript{\rm 1} Institute of Information Engineering, Chinese Academy of Sciences  \\
	\textsuperscript{\rm 2} School of Cyber Security, University of Chinese Academy of Sciences \\
	\{yuanchunyuan,maqianwen,zhouwei,hanjizhong,husonglin\}@iie.ac.cn
}

\begin{document}

\maketitle
\begin{abstract}
The\let\thefootnote\relax\footnotetext{* Corresponding author.} dissemination of fake news significantly affects personal reputation and public trust. Recently, fake news detection has attracted tremendous attention, and previous studies mainly focused on finding clues from news content or diffusion path. However, the required features of previous models are often unavailable or insufficient in early detection scenarios, resulting in poor performance. Thus, early fake news detection remains a tough challenge. Intuitively, the news from trusted and authoritative sources or shared by many users with a good reputation is more reliable than other news. Using the credibility of publishers and users as prior weakly supervised information, we can quickly locate fake news in massive news and detect them in the early stages of dissemination.

In this paper, we propose a novel \textbf{S}tructure-aware \textbf{M}ulti-head \textbf{A}ttention \textbf{N}etwork (SMAN), which combines the news content, publishing, and reposting relations of publishers and users, to jointly optimize the fake news detection and credibility prediction tasks. In this way, we can explicitly exploit the credibility of publishers and users for early fake news detection. We conducted experiments on three real-world datasets, and the results show that SMAN can detect fake news in 4 hours with an accuracy of over 91\%, which is much faster than the state-of-the-art models. The source code and dataset can be available at https://github.com/chunyuanY/FakeNewsDetection.
\end{abstract}

\section{Introduction}

\blfootnote{
	%
	%
	%
	%
	%
	%
	This work is licensed under a Creative Commons 
	Attribution 4.0 International License.
	License details:
	\url{http://creativecommons.org/licenses/by/4.0/}.
}

The widespread dissemination of fake news has lead to a significant influence on personal fame, public trust, and security. For example, spreading misinformation, such as ``Asians are more vulnerable to novel coronavirus''~\footnote{https://www.thestar.com.my/news/regional/2020/03/11/myth-busters-10-common-rumours-about-covid-19} about COVID-19 has very serious repercussions, making people ignore the harmfulness of the virus and directly affecting public health. Research has shown that misinformation spreads faster, farther, deeper, and more widely than true information~\cite{vosoughi2018spread}. Therefore, fake news detection on social media has attracted tremendous attention recently in both research and industrial fields. 

Early research on fake news detection mainly focused on the design of effective features from various sources, including textual content, user profiling data, and news diffusion patterns. Linguistic features, such as writing styles and sensational headlines~\cite{Kwon_2013}, lexical and syntactic analysis~\cite{potthast2017stylometric}, have been explored to separate fake news from true news. Apart from linguistic features, some studies also proposed a series of user-based features~\cite{Castillo_2011,shu2018understanding}, and temporal features~\cite{Kwon_2013} about the news diffusion. However, these feature-based methods are very time-consuming, biased, and require a lot of labor to design. Besides, these features are easily manipulated by users.

To solve the above problems, many recent studies~\cite{ma2016detecting,Yu_2017,guo2018rumor,shu2019beyond,rumor_yuan_2019} apply various neural networks to automatically learn high-level representations for fake news detection. For example, recurrent neural network (RNN)~\cite{ma2016detecting}, convolutional neural network (CNN)~\cite{Yu_2017}, matrix factorization~\cite{shu2019beyond} and graph neural network~\cite{rumor_yuan_2019} are applied to learn the representation of content and diffusion graph of news. These methods only apply more types of information for fake news detection, but paying little attention to early detection. Moreover, these models can only detect fake news in consideration of all or a fixed proportion of repost information, while in practice they cannot detect fake news in the early stage of news propagation~\cite{song2018ced}. Some studies~\cite{liu2018early,song2018ced,zhou2019early} explore to detect fake news early by relying on a minimum number of posts. The main limitation of these methods is that they ignore the importance of publishers' and users' credibility for the early detection of fake news.

When we humans see a piece of breaking news, we firstly may use common sense to judge whether there are factual errors in it. At the same time, we will also consider the reputation of the publishers and reposted users. People tend to believe the news from a trusted and authoritative source or the news shared by lots of users with a good reputation. If the publisher is reliable, we tend to believe this news. On the other hand, if the news is reposted by many low-reputation users in a short period, it may be that some spammers tried to heat up on the news~\cite{chen2015opinion,vosoughi2018spread}, resulting in lower credibility of the news. 

Inspired by the above observation, we explicitly take the credibility of publishers and users as supervised information, and model fake news detection as a multi-task classification task. We can annotate a small part of publishers and users by their historical publishing and reposting behaviors. Although the credibility of publishers and users does not always provide correct information, they are necessary complementary supervised information for fake news detection. To make the credibility information generalized to other unannotated users, we construct a heterogeneous graph to build the connections of publishers, news, and users. Through a graph-based encoding algorithm, every node in the graph will be influenced by the credibility of publishers and users.

In this paper, we address the following challenges: (1) How to fully encode the heterogeneous graph structure and news content; and (2) How to explicitly utilize the credibility of publishers and users for facilitating early detection of fake news. To tackle the above challenges, we propose a novel structure-aware multi-head attention network for the early detection of fake news. Firstly, we design a structure-aware multi-head attention module to learn the structure of the publishing graph and produce the publisher representations for the credibility prediction of publishers. Then, we apply the structure-aware multi-head attention module to encode the diffusion graph of the news among users and generate user representations for the credibility prediction of users. Finally, we apply a convolutional neural network to map the news text from word embedding to semantic space and utilize the fusion attention module to combine the news, publisher, and user representations for early fake news detection. 

The contributions of this paper can be summarized as follows:
\begin{itemize}
	\item We propose a novel strategy that explicitly takes the credibility of publishers and users as weakly supervised information for facilitating early detection of fake news.
	
	\item We provide a principled way to jointly utilize the credibility of publishers and users, and the heterogeneous graph for credibility prediction and fake news detection.
	
	\item We conduct extensive experiments on three real-world datasets. Experimental results show that our model achieves significant improvement over state-of-the-art models on both fake news detection and early detection tasks.
\end{itemize}

\section{Related Work}  \label{related_works_section}

\subsection{Feature-based Methods}
Early studies in fake news detection concentrate on designing some good features for separating fake news from true news. These features are mainly extracted from text content or users' profile information. Linguistic patterns, such as special characters and keywords~\cite{Castillo_2011}, writing styles and sensational headlines~\cite{Kwon_2013}, lexical and syntactic features~\cite{feng2012syntactic,potthast2017stylometric}, temporal-linguistic features~\cite{Ma_2015,zhao2015enquiring}, have been explored to detect fake news. Apart from linguistic features, some studies also proposed a series of user-based features~\cite{Castillo_2011,Yang_2012}, e.g. the number of fans, registration age, and genders~\cite{Castillo_2011} to find clues for fake news detection.

However, the language used in social media is highly informal and ungrammatical, which makes traditional natural language processing techniques hard to effectively learn semantic information from news content. Second, designing effective functions is often time-consuming and relies heavily on expert knowledge in specific fields. Some features are often unavailable or inadequate in the early stage of news propagation.

\subsection{Deep Learning Methods}
Recurrent neural network (RNN)~\cite{ma2016detecting}, convolutional neural network (CNN)~\cite{Yu_2017} and graph neural network~\cite{rumor_yuan_2019} have been imported to learn the representations from news content or diffusion graph. Some studies also combine news content and users' response, such as conflicting viewpoints~\cite{jin2016news}, topics~\cite{guo2018rumor}, or stance~\cite{Bhatt_2018,li2019rumor}, to find clues by neural networks for fake news detection. These methods only apply more types of information for fake news detection, but paying little attention to early detection. 

Recently, some studies~\cite{liu2018early,song2018ced,shu2019beyond,zhou2019early} have proposed some methods to detect fake news at the early stage of propagation. However, these methods ignored the importance of publishers' and users' credibility for the early detection of fake news. Different from these studies, our method explicitly takes the credibility of publishers and users as weakly supervised information for facilitating fake news detection. We propose a novel deep learning model to simultaneously optimize the fake news detection task and users' credibility prediction task.

\section{Problem Formulation} \label{formulation_section}
Let $\mathcal{N}=\left\{ m_{1},m_{2},\ldots m_{|\mathcal{N}|} \right\} $ be the set of news. Each news $m_j$ has one publisher at least and $K$ users $\{R_1, R_2, \ldots, R_K \}$ to repost it at most. The publisher-news relations form a publishing graph $\mathcal{G}(V_p, E)$. The publisher-user relations form a diffusion graph $\mathcal{G}(V_u, E)$. In the diffusion graph of news, we regard users who repost the news as the neighbor nodes of the publisher. We use $|P|$, $|N|$, and $|U|$ to denote the amount of publishers, news, and users respectively.

For fake news detection task, our target is to learn a function $p(c| m_j, \mathcal{P}, \mathcal{N}, \mathcal{U}; \mathcal{\theta}_3)$ to predict whether a piece of news is fake or not. $c$ is class label of the news and $\mathcal{\theta}_3$ represents all parameters of the model. 

In this paper, we design a credibility prediction subtask to explicitly utilize the users' or publishers' credibility information for fake news detection. For credibility prediction task, our goal is to learn a function $p(c|\mathcal{G}(V_p, E), \mathcal{P}; \mathcal{\theta}_1)$ or $p(c|\mathcal{G}(V_u, E), \mathcal{U}; \mathcal{\theta}_2)$ to predict the credit scores of publishers or users by publishing graph or diffusion graph.

\section{The Proposed Framework} \label{model_section}
\begin{figure}[htbp]
	\centering
	\includegraphics[scale=0.7]{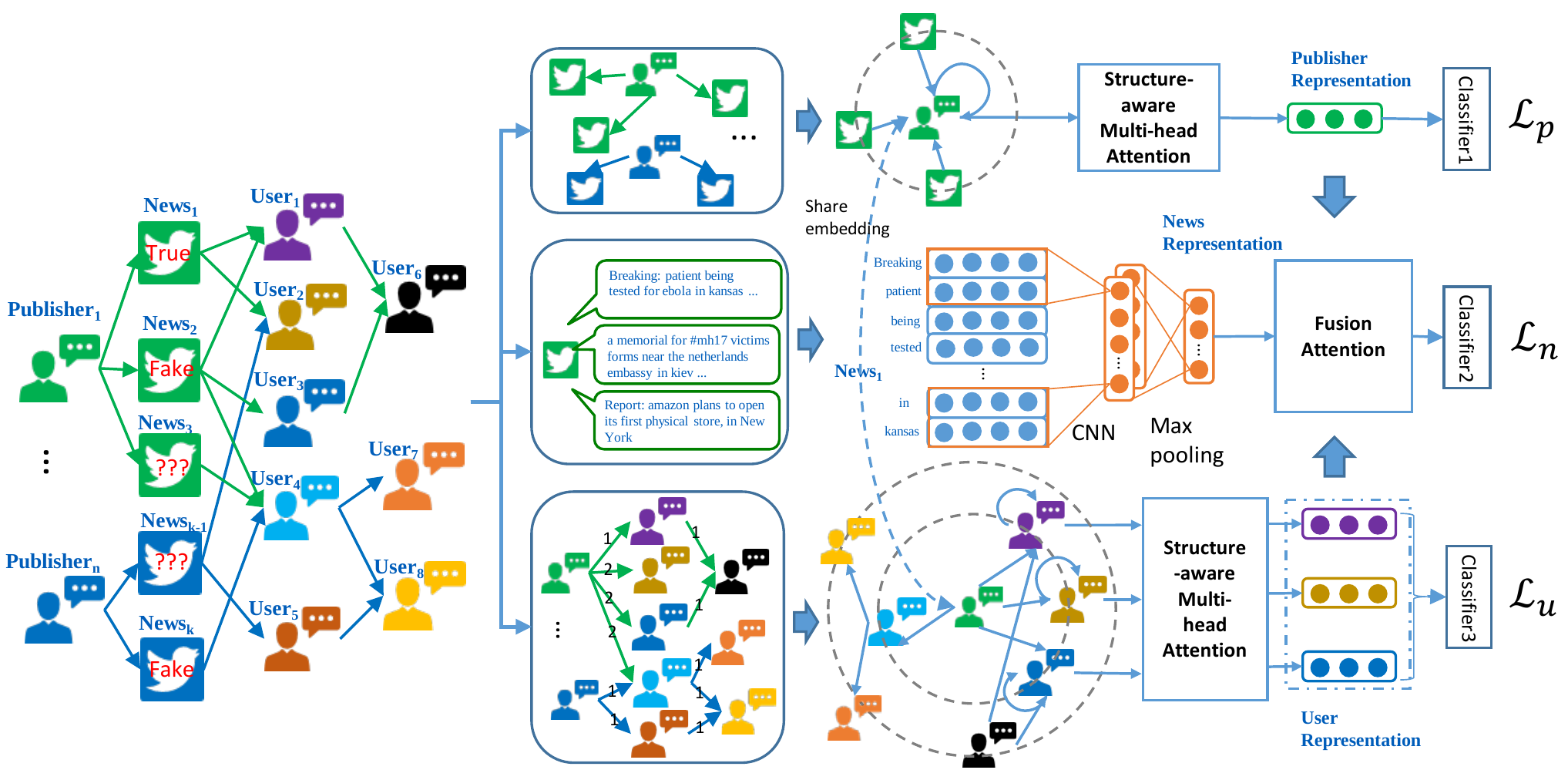}
	\vspace{-1\baselineskip}
	\caption{The architecture of the proposed fake news detection model.}
	\label{model}
	\vspace{-1\baselineskip}
\end{figure}

The proposed framework consists of three major components: (1) publisher credibility prediction; (2) user credibility prediction;  and (3) fake news classification. Figure~\ref{model} illustrates the architecture of the proposed model.

\subsection{Publisher Credibility Prediction}
In recent years, the multi-head attention mechanism~\cite{vaswani2017attention} shows the superior ability to learn the semantic representations of documents in the natural language process, which inspires us to extend it to learn node representations for graph representation learning. In this paper, we extend the Multi-head Attention~\cite{vaswani2017attention} as a structure-aware multi-head attention module to encode the structure of the graph and learn the node representation from the publishing graph.

The structure-aware multi-head attention module has three input items: the query item, the key item and the value item, namely $Q \in \mathbb{R}^{n_q \times d}$, $K \in \mathbb{R}^{n_k \times d}$, and $V \in \mathbb{R}^{n_v \times d}$ respectively, where $n_q$, $n_k$, and $n_v$ denote the number of nodes in each item, and $d$ is the dimensionality of the node embedding. The attention module first takes each node in the query to attend to all nodes in the key item via a dot-product attention unit. But in fact, it is impossible for each node to establish connections with all nodes in the social graph. Thus, we encode the adjacent relations of the graph structure into the attention module. The adjacency matrix $\mathbf{A}^{pn} \in \mathbb{R}^{|P| \times |N|}$, whose element $\mathbf{A}_{ij}^{pn}$ denotes that publisher $i$ deliver a piece of news $j$. Finally, we apply those attention weights upon the value item: 
\begin{equation}
	\begin{split}
		& \mathbf{Attention}(\mathbf{Q}, \mathbf{K}, \mathbf{V})_h = \mathbf{softmax}\left(\frac{\mathbf{Q} \mathbf{W}_h \mathbf{K}^T }{\sqrt{d}}  \odot (\mathbf{D}^p)^{-\frac{1}{2}} \mathbf{A}^{pn} (\mathbf{D}^n)^{-\frac{1}{2}} \right) \mathbf{V} \,, \\
	\end{split}
\end{equation}
where $\mathbf{W}_h \in \mathbb{R}^{d \times d}$ is a transformation matrix. $(\mathbf{D}^p)_{ii} = \sum_{j} \mathbf{A}^{pn}_{ij}$ and $(\mathbf{D}^n)_{jj} = \sum_{i} \mathbf{A}^{pn}_{ij}$ are diagonal matrices, which are applied to normalize the adjacency matrix $\mathbf{A}^{pn}$. $\odot$ denotes element-wise product.

The entries of $\textbf{V}$ are then linearly combined with the weights to form a new representation of $\textbf{Q}$. In this way, the structure-aware attention module can capture relations across query nodes and key nodes, and further use the relations to aggregate embeddings in the query to produce new node representations. We usually let $\textbf{K} = \textbf{V}$. Therefore, every node in $\textbf{Q}$ is represented by its most similar nodes in $\textbf{V}$. 

For each head of attention captures relations among $\textbf{Q}$, $\textbf{K}$, and $\textbf{V}$ from one aspect, we expand one head attention to multi-head schema: $\textbf{Q}$, $\textbf{K}$, and $\textbf{V}$ are dispensed to $h$ heads. Specifically, $\forall h \in [1, H] $ the output of head $h$ is given by following formulation:
\begin{equation}
	\begin{split}
		& \mathbf{Z}_h = \textbf{Attention}(\textbf{P}, \textbf{N}, \textbf{N})_h  \,,   h \in [1, H] \\
	\end{split}
\end{equation}
where $\mathbf{P} \in \mathbb{R}^{|P| \times d}$ is the publishers' embeddings and $\mathbf{N} \in \mathbb{R}^{|N| \times d}$  is the news embeddings. $H$ is the amount of heads in attention module. Every publisher and news is transformed into a $d$-dimensional embedding by their id and the vector is initialized by normal distribution~\cite{glorot2010understanding}. 

Then, the output features of multi-head attention are concatenated together and a fully-connected layer is applied to transform it as final output, which is formalized as:
\begin{equation}
	\begin{split}
		& \mathbf{\tilde{P}} = \mathbf{ELU}\left( \left[\mathbf{Z}_1; \mathbf{Z}_2; \ldots; \mathbf{Z}_H \right] \mathbf{W}_o \right) + \textbf{P} \,, \\
	\end{split}
\end{equation}
where $\mathbf{W}_o \in \mathbb{R}^{Hd \times d} $ is a linear transformation matrix and $\mathbf{ELU}(x)$ is an activation function.

We obtain publishers' representations $\mathbf{\tilde{P}} \in \mathbb{R}^{|P| \times d}$ after above procedure. Finally, we use these features to predict the publishers' credibility, which can be formulated as follows:
\begin{equation}
	\begin{split}
		& p_i(c|\mathcal{G}(V_p, E), \mathcal{P}; \mathcal{\theta}_1) = \mathop{\textbf{softmax}}(\mathbf{\tilde{P}}_i \mathbf{W}_p + \mathbf{b}_p)  \,,   \\
	\end{split}
\end{equation}
where $\mathbf{b}_p$ is a bias term, and $\mathbf{W_p} \in \mathbb{R}^{d \times |c|}$ and $|c|$ is the total levels of credibility. The credit scores have three levels ($|c|=3$): ``unreliable'', ``uncertain'', and ``reliable''. The annotation of credibility will be introduced in Section~\ref{datasetintro}. Finally, the publisher credibility prediction task can be transformed into a classification task.

We apply the cross-entropy loss as the optimization objective:
\begin{equation}
	\begin{split}
		& \mathcal{L}_p = -\sum_{i=1}^{|P|} y^{(p)}_i\log p_i(c| \mathcal{G}(V_p, E), \mathcal{P}; \mathcal{\theta}_1) + \frac{\lambda}{2} ||\mathcal{\theta}_1||^2_2 \,, \\
	\end{split}
	\label{loss1}
\end{equation}
where $y^{(p)}_i$ is the true credibility of publisher $i$ and $\mathcal{\theta}_1$ denotes all parameters need to be trained in this subtask. We apply $\ell_2$ regularization on all parameters of the model to overcome overfitting problem. $\lambda$ is a regularization factor. 

\subsection{User Credibility Prediction}
Same as publisher credibility prediction task, we apply user credibility as weakly supervised information to facilitate fake news detection. Firstly, we construct the diffusion graph of news $\mathcal{G}(V_u, E)$, which records how news propagated from publishers to other users. The nodes $V_u$ of the graph belongs to the user set and the edges denote the diffusion traces.

Suppose that every news will be reposted by $K$ different users at most. We use matrix $\mathbf{R} \in \mathbb{R}^{|U| \times K} $ to denote the user ids who had reposted the news before. `0' is padded at the start of the matrix $\mathbf{R}$ when the amount of reposted users is less than $K$. We still apply structure-aware multi-head attention to learn the user node representation from the diffusion graph. The attention unit is defined as follows:
\begin{equation}
	\begin{split}
		& \mathbf{Attention}(\mathbf{Q}, \mathbf{K}, \mathbf{V})_h = \mathbf{softmax}\left(\frac{\mathbf{Q} \mathbf{W}_h \mathbf{K}^T }{\sqrt{d}}  \odot (\mathbf{D}^u)^{-\frac{1}{2}}\mathbf{A}^{uu} (\mathbf{D}^u)^{-\frac{1}{2}} \right) \mathbf{V} \,, \\
	\end{split}
	\label{user_attention}
\end{equation}
where $\mathbf{W}_h \in \mathbb{R}^{d \times d}$ is a transformation matrix. $(\mathbf{D}^u)_{ii} = \sum_{j} \mathbf{A}^{uu}_{ij}$ is a diagonal matrix, which is used to normalize the adjacency matrix $\mathbf{A}^{uu}$. The complete computation process is shown in Algorithm~\ref{alg_global}.

\begin{algorithm}[!htb] 
	\caption{The diffusion graph encoding algorithm.}
	\small
	\label{alg_global} 
	\KwIn{ \\
		\quad 1. Adjacency matrix $\mathbf{A}^{uu}$ of the diffusion graph $\mathcal{G}(V_u, E)$; \\
		\quad 2. User embeddings $\mathbf{U} \in \mathbb{R}^{|U| \times d}$;  \\
		\quad 3. $\mathbf{Lookup}(\cdot)$ that can extract user embedding from $\mathbf{U}$ by user id.   \\
		\quad 4. Weight matrices $\mathbf{W}_o \in \mathbb{R}^{Hd \times d}$, $\mathbf{W}_h \in \mathbb{R}^{d \times d}$ and $h \in [1, 2, \ldots, H]$; \\
		\quad 5. User ids matrix $R \in \mathbb{R}^{|U| \times K} $. \\
	}
	
	\KwOut{
		User representations $\mathbf{\tilde{R}}$.
	}
	
	\For{$j \in [1,2, \ldots, K] $}
	{
		\For{$h \in [1,2, \ldots, H]$}
		{
			$\mathbf{R}_j = \mathbf{Lookup}(R_j) $  \;  
			Calculate $\mathbf{Z}_h = \textbf{Attention}(\mathbf{R}_j, \mathbf{U}, \mathbf{U})_h$ by Equation~(\ref{user_attention}) \;  
		}
		$\mathbf{\tilde{R}}_j = \mathbf{ELU}\left( \left[\mathbf{Z}_1; \mathbf{Z}_2; \ldots; \mathbf{Z}_H \right] \mathbf{W}_o \right) + \textbf{R}_j$
	}
	\Return $\mathbf{\tilde{R}} = [\mathbf{\tilde{R}}_1; \mathbf{\tilde{R}}_2; \ldots ; \mathbf{\tilde{R}}_K] $
\end{algorithm}

To learn abundant representations from different reposting relations, we extend structure-aware attention to employ a multi-head paradigm. Specifically, $H$ independent attention units execute the transformation of Equation~\ref{user_attention}, and then their features are concatenated, resulting in the user representations.

Finally, we use these users' representations $\mathbf{\tilde{R}} \in \mathbb{R}^{|U| \times K \times d}$ to predict the users' credibility scores, which can be formulated as follows:
\begin{equation}
	\begin{split}
		& p_{ij}(c|\mathcal{G}(V_u, E), \mathcal{U}; \mathcal{\theta}_2) = \mathop{\textbf{softmax}}(\mathbf{\tilde{R}}_{ij} \mathbf{W}_r + \mathbf{b}_r)  \,,   \\
	\end{split}
\end{equation}
where $i \in [1, \ldots, |U| ]$ and $j \in [1, \ldots, K ]$. $\mathbf{W}_r \in \mathbb{R}^{d \times |c|}$ is a trainable matrix and $|c|$ is the levels of credibility. $\mathbf{b}_r \in \mathbb{R}^{|c|}$ is a bias term.

The credit scores of users are annotated in the same way as the credit scores of publishers. We apply the cross-entropy loss as the optimization function:
\begin{equation}
	\begin{split}
		& \mathcal{L}_u = -\sum_{i=1}^{|U|} \sum_{j=1}^{K} y^{(u)}_{ij}\log p_{ij}(c| \mathcal{G}(V_u, E), \mathcal{U}; \mathcal{\theta}_2)  + \frac{\lambda}{2} ||\mathcal{\theta}_2||^2_2  \,, \\
	\end{split}
	\label{loss2}
\end{equation}
where $y^{(u)}_{ij}$ is the credibility of user $u_{ij}$ and $\mathcal{\theta}_2$ denotes all parameters needed to be trained in this subtask.

\subsection{Fake News Classification}
For the fake news classification, we combine news with publishing and diffusion graph to more comprehensively capture the differences in the content and diffusion mode of true and false news.

\subsubsection{News Content Representation}
There have been many natural language processing models that can be used to learn the text representation from word sequence embeddings, such as CNN~\cite{kim2014convolutional,kalchbrenner2014convolutional} and RNN~\cite{tai2015improved,yang2016hierarchical}. For a fair comparison, we also apply CNN~\cite{kim2014convolutional} as the basic component to learn the representation of news, which is the same as paper~\cite{rumor_yuan_2019}.

\subsubsection{Fusion Attention Unit}
After the above process, we have obtained news content representation $\mathbf{m}_j \in \mathbb{R}^{3d}$ for news $m_j$ from word embeddings by CNN. Then, we will introduce how to fuse the publisher, user, and content representations for classification.

Firstly, we find publisher id $p_i$ from the publishing and diffusion graph by news id $m_j$. Then, we look up publisher representation $\mathbf{\tilde{P}}_i \in \mathbb{R}^{d}$ from all publisher representations table $\mathbf{\tilde{P}}$ by publisher id $p_i$. And by the same way, we look up user representations $\mathbf{\tilde{R}}_i \in \mathbb{R}^{K \times d}$ from all user representations table $\mathbf{\tilde{R}}$ by publisher id $p_i$. $\mathbf{\tilde{R}}_i$ denotes $K$ different users who had reposted the news $m_j$. 

We aggregate the reposted user embeddings $\mathbf{\tilde{R}}_i \in \mathbb{R}^{K \times d}$ by an attention module: 
\begin{equation}
	\mathbf{R}' = \sum_{k=1}^{K} \alpha_k \mathbf{\tilde{R}}_k  \,, \quad  \alpha =  \mathop{\textbf{softmax}}( \mathbf{N}_j \mathbf{\tilde{R}}_i^T  )  \,,  \\
\end{equation}
where $\mathbf{N}_j \in \mathbb{R}^{1 \times d}$ is the embedding of news $m_j$ looked up from the news embeddings table $\mathbf{N}$.

Then, we fuse the publisher representation and user combined representation by a heuristic method:
\begin{equation}
	\begin{split}
		& \mathbf{\tilde{m}}_j =  [\mathbf{\tilde{P}}; \mathbf{R}'; \mathbf{\tilde{P}} \odot \mathbf{R}'; \mathbf{\tilde{P}} - \mathbf{R}'] \mathbf{W}_F + \mathbf{b}_F \,, \\
	\end{split}
\end{equation}
where $\mathbf{W}_F \in \mathbb{R}^{4d \times d}$ is transformation matrix and $\mathbf{b}_F \in \mathbb{R}^d$ is a bias term.

News content representation captures the semantic difference between fake and true news. $\mathbf{\tilde{m}}_j$ captures the differences between fake and true news from the diffusion graph. Both representations are important for fake news detection, thus they are concatenated as final features. A fully-connected layer is applied to project the final representation into the target space of classes probability:
\begin{equation}
	\begin{split}
		& p(c| m_j, \mathcal{P}, \mathcal{N}, \mathcal{U}; \mathcal{\theta}_3) = \mathop{\textbf{softmax}}([\mathbf{m}_j; \mathbf{\tilde{m}}_j] \mathbf{W}_m + b)  \,,   \\
	\end{split}
\end{equation}
where $\mathbf{W}_m \in \mathbb{R}^{4d \times |c|}$ is a transformation matrix and $b \in \mathbb{R}$ is a bias term. 

Finally, the cross-entropy loss is used as the optimization objective function for fake news detection: 
\begin{equation}
	\begin{split}
		& \mathcal{L}_n = -\sum_{j=1}^{|N|}y^{(n)}_j\log p(c| m_j, \mathcal{P}, \mathcal{N}, \mathcal{U}; \mathcal{\theta}_3) + \frac{\lambda}{2} ||\mathcal{\theta}_3||^2_2  \,, \\
	\end{split}
	\label{loss3}
\end{equation}
where $y^{(n)}_j$ is the gold class probability of news $m_j$.

For simultaneously optimize the credibility prediction task and fake news detection task, we combine all these optimization objective as follows:
\begin{equation}
	\begin{split}
		& \mathcal{L}\left(c| \mathcal{G}(V_p, E), \mathcal{G}(V_u, E), \mathcal{N}; \mathcal{\theta} \right) = \mathcal{L}_p + \mathcal{L}_u + \mathcal{L}_n \,,  \\
	\end{split}
\end{equation}
where $\mathcal{\theta} = \{ \mathcal{\theta}_1, \mathcal{\theta}_2, \mathcal{\theta}_3 \}$ represents all parameters of the model SMAN.

\section{Experiments} \label{experiments}
In this section, we introduce the experiments to evaluate the effectiveness of SMAN. Specifically, we aim to answer the following evaluation questions:
\begin{itemize}
	\item EQ1: Can SMAN improve fake news classification performance by jointly optimizing the fake news detection task and publishers' and users' credibility prediction task?
	\item EQ2: How effective are publishers' and users' credibility prediction tasks, respectively, in improving the detection performance of SMAN?
	\item EQ3: Can SMAN improve the performance of fake news early detection task?
\end{itemize}

\subsection{Datasets}  \label{datasetintro}
We evaluate SMAN on three real-world datasets: Twitter15~\cite{ma2017detect}, Twitter16~\cite{ma2017detect}, and Weibo~\cite{ma2016detecting}. Table~\ref{tab1} shows the statistics of the three datasets.

\begin{table}[!htbp]
	\vspace{-1\baselineskip}
	\small
	\centering
	\caption{Dataset statistics. The label ``true news'' denotes a microblog that debunks the fake news.}
	\setlength{\tabcolsep}{0.4mm}{
		\begin{tabular}{|c|c|c|c|c|c|c|c|}
			\hline
			\textbf{Statistic} &\# news &\# non-fake news (NR) &\# fake news (FR) & \# unverified news (UR) & \# true news (TR) & \# users & \# retweets \\
			\hline
			\textbf{Twitter15} & 1490  & 374   & 370   & 374   & 372  & 276,663  & 331,612  \\
			\hline
			\textbf{Twitter16} & 818   & 205   & 205   & 203   & 205  & 173,487  &  204,820 \\
			\hline
			\textbf{Weibo}     & 4664  & 2351  & 2313  &  0    &  0  & 2,746,818  &   3,805,656 \\
			\hline
		\end{tabular}
	}
	\label{tab1}
	\vspace{-0.5\baselineskip}
\end{table}

For a fair comparison, we use the train, validation, and test set that is split by~\cite{rumor_yuan_2019}, where 10\% samples as the validation dataset, and split the rest for training and test set with a ratio of 3:1. 

The credit scores of publishers and users in these three datasets are annotated according to the training set. In this paper, we have defined three levels of credibility for publishers and users: (1) ``0'' means ``reliable'' (the publisher has never delivered fake or unverified news); (2) ``1'' means ``uncertain'' (the publisher not only delivers true news, but also publishes false news); (3) ``2'' means ``unreliable'' (publishers always publish false news and unverified news, but never publish true news).

\subsection{Baseline Models}  \label{baselinemodel}
We compare our model with a series of fake news detection methods as follows: 

(1) Feature-based methods: \textbf{DTC}~\cite{Castillo_2011}: A decision tree-based model that utilizes a combination of news characteristics. \textbf{SVM-RBF}~\cite{Yang_2012}: An SVM model with RBF kernel that utilize the news features. \textbf{SVM-TS}~\cite{Ma_2015}: An SVM model that utilizes time-series to model the variation of news characteristics. \textbf{DTR}~\cite{Zhao_2015}: A decision-tree-based method for detecting fake news through enquiry phrases. \textbf{RFC}~\cite{Kwon_2017}: A random forest classifier that utilizes user, linguistic and structure features. \textbf{cPTK}~\cite{ma2017detect}: An SVM classifier with a propagation tree kernel that detects fake news by learning temporal-structure patterns.

(2) Deep Learning methods: \textbf{GRU}~\cite{ma2016detecting}: A RNN-based model that learns temporal-linguistic patterns from user comments. \textbf{RvNN}~\cite{ma2018rumor}: A bottom-up and a top-down tree-structured model based on recursive neural networks for fake news detection on Twitter. \textbf{PPC}~\cite{liu2018early}: A model that detects fake news through propagation path classification with a combination of recurrent and convolutional networks. \textbf{GLAN}~\cite{rumor_yuan_2019}: A model that jointly encodes the local semantic and global structure of the diffusion graph.

\subsection{Evaluation Metrics and Parameter Settings}  \label{parametersettings}
Same as previous studies~\cite{liu2018early,ma2018rumor,rumor_yuan_2019}, we also adopt accuracy, precision, recall and F1 score as the evaluation metrics. The parameters of SMAN are updated by Adam algorithm~\cite{sashank2018iclr} with default parameters. All word embeddings of the model are initialized with the 300-dimensional word vectors, which is released by \cite{rumor_yuan_2019}. The convolutional kernel size is set to (3, 4, 5) with 100 kernels for each kind of size. The number of heads in structure-aware multi-head attention $H$ is chosen from $\{1,2,3, \ldots, 11, 12 \}$ and is set to 7. The $\lambda$ in Equation~(\ref{loss1}),~(\ref{loss2}),~(\ref{loss3}) is chosen from $\{1e^{-8}, 1e^{-7}, \ldots, 1e^{-2} \}$ and is set to $1e^{-6}$.

\subsection{Results and Analysis}
To answer EQ1, we compare SMAN with baselines introduced in Section~\ref{baselinemodel} for fake news classification. The experimental results of all baseline methods are shown in Table~\ref{exp_results_on_twitter15}, \ref{exp_results_on_twitter16}, and \ref{exp_results_on_weibo}. For fair comparison, the performance of baselines is directly cited from previous studies~\cite{ma2018rumor,liu2018early,rumor_yuan_2019}. The \textbf{GLAN} model is the state-of-the-art method when submitting this paper.

\begin{table}[!htbp]
	\vspace{-1\baselineskip}
	\small
	\begin{minipage}[t]{0.49\linewidth}
		\centering
		\caption{Experimental results on Twitter15 dataset. }
		\setlength{\tabcolsep}{1mm}{
			\begin{tabular}{|c|c|cccc|}
				\hline
				\multirow{2}{*}{Method} & \multirow{2}{*}{Accuracy} & NR & FR & TR & UR \\ \cline{3-6}
				&  & $F_1$ & $F_1$ & $F_1$ & $F_1$ \\
				\hline
				DTR    & 0.409     & 0.501     & 0.311 & 0.364 & 0.473  \\
				\hline
				DTC    & 0.454     & 0.733     & 0.355 & 0.317 & 0.415  \\
				\hline
				RFC    & 0.565     & 0.810     & 0.422 & 0.401 & 0.543  \\
				\hline
				SVM-RBF    & 0.318     & 0.455     & 0.037 & 0.218 & 0.225  \\
				\hline
				SVM-TS    & 0.544     & 0.796     & 0.472 & 0.404 & 0.483  \\
				\hline
				cPTK    & 0.750     & 0.804     & 0.698 & 0.765 & 0.733  \\
				\hline
				GRU    & 0.646     & 0.792     & 0.574 & 0.608 & 0.592  \\
				\hline
				RvNN    & 0.723     & 0.682     & 0.758 & 0.821 & 0.654  \\
				\hline
				PPC    & 0.842     & 0.811     & 0.875 & 0.818 & 0.790  \\
				\hline
				GLAN    & 0.905    & \textbf{0.924}     & 0.917 & 0.852  & 0.927  \\
				\hline
				SMAN    & \textbf{0.929}  & 0.922  & \textbf{0.945}  & \textbf{0.915}  & \textbf{0.933} \\
				\hline
			\end{tabular}
			\label{exp_results_on_twitter15}
		}
	\end{minipage} 
	\begin{minipage}[t]{0.49\linewidth}  
		\centering
		\caption{
			Experimental results on Twitter16 dataset. 
		}
		\setlength{\tabcolsep}{1mm}{
			\begin{tabular}{|c|c|cccc|}
				\hline
				\multirow{2}{*}{Method} & \multirow{2}{*}{Accuracy} & NR & FR & TR & UR \\ \cline{3-6}
				&  & $F_1$ & $F_1$ & $F_1$ & $F_1$ \\
				
				\hline
				DTR    & 0.414     & 0.394     & 0.273 & 0.630 & 0.344  \\
				\hline
				DTC    & 0.465     & 0.643 & 0.393& 0.419 & 0.403  \\
				\hline
				RFC    & 0.585     & 0.752 & 0.415 & 0.547 & 0.563  \\
				\hline
				SVM-RBF    & 0.321     & 0.423 & 0.085 & 0.419 & 0.037 \\
				\hline
				SVM-TS    & 0.574     & 0.755 & 0.420 & 0.571 & 0.526  \\
				\hline
				cPTK    & 0.732     & 0.740 & 0.709 & 0.836 & 0.686  \\
				\hline
				GRU    & 0.633     & 0.772 & 0.489 & 0.686 & 0.593  \\
				\hline
				RvNN    & 0.737     & 0.662     & 0.743 & 0.835 & 0.708  \\
				\hline
				PPC    & 0.863     & 0.820 & 0.898 & 0.843 & 0.837  \\
				\hline
				GLAN    & 0.902     & 0.921     & 0.869 & 0.847 & 0.968  \\
				\hline
				SMAN    & \textbf{0.935}     & \textbf{0.946}     & \textbf{0.920} & \textbf{0.894} & \textbf{0.979}  \\
				\hline
				
			\end{tabular}
			\label{exp_results_on_twitter16}
		}
	\end{minipage}
\end{table}

\begin{table}[!htbp]
	\vspace{-1.5\baselineskip}
	\centering
	\small
	\caption{Fake news detection results on Weibo dataset.}
	\setlength{\tabcolsep}{2mm}{
		\begin{tabular}{|c|c|ccc|ccc|}
			\hline
			\multirow{2}{*}{Method} & \multirow{2}{*}{Acc} & \multicolumn{3}{c|}{NR} & \multicolumn{3}{c|}{FR}   \\ \cline{3-8}
			&  & Precision & Recall & $F_1$ & Precision & Recall & $F_1$ \\
			\hline
			DTR      &  0.732    & 0.726 & 0.749 & 0.737   & 0.738 & 0.715 & 0.726 \\
			\hline
			DTC      &  0.831    & 0.815 & 0.847 & 0.830   & 0.847 & 0.815 & 0.831     \\
			\hline
			RFC      &  0.849    & 0.947 & 0.739 & 0.830   & 0.786 & 0.959 & 0.864   \\
			\hline
			SVM-RBF  &  0.818    & 0.815 & 0.824 & 0.819   & 0.822 & 0.812 & 0.817  \\
			\hline
			SVM-TS   &  0.857    & 0.878 & 0.830 & 0.857   & 0.839 & 0.885 & 0.861   \\
			\hline
			GRU      &  0.910    & \textbf{0.952} & 0.864 & 0.906   & 0.876 & 0.956 & 0.914     \\
			\hline
			PPC      &  0.921    & 0.949 & 0.889 & 0.918   & 0.896 & \textbf{0.962} & 0.923  \\
			\hline
			GLAN     &  0.946    & 0.949 & 0.943 & 0.946   & 0.943 & 0.948 & 0.945    \\
			\hline
			SMAN     &  \textbf{0.951}    & 0.937 & \textbf{0.967} & \textbf{0.952}  & \textbf{0.967} & 0.936 & \textbf{0.951}    \\
			\hline
		\end{tabular}

	}
	\label{exp_results_on_weibo}
\end{table}

\noindent We bold the best performance of each column in all tables. From the tables, we can observe that:

(1) Methods based on manually designed features (DTR, DTC, RFC, SVM-RBF, cPTK, and SVM-TS) have poorer performance. It indicates 1) hand-crafted features cannot effectively encode semantic information of news content; 2) these methods cannot perform deep feature interaction; thus unable to fully learn the difference between fake and true news. 

(2) Deep learning methods (GRU, RvNN, PPC, and GLAN) significantly outperform conventional classifiers that using manually designed features. This observation indicates deep learning models can learn better semantic representations and perform better feature interactions. We can also observe that GLAN is more effective than RvNN and PPC because it can deeply integrate local semantic and global diffusion structure for fake news detection.

(3) SMAN achieves significant improvement compared with GLAN. Different from GLAN, SMAN not only optimizes the fake news detection task but also tries to predict the credibility of publishers and users. The results show that the credibility of publishers and users is critical for learning the differences between fake and true news.

\subsection{Ablation Study}
To answer EQ2, we further perform some ablation studies over the different modules of SMAN. The experimental results are presented in Table~\ref{ablation_results}.

\begin{table}[!htb]
	\vspace{-1\baselineskip}
	\small
	\centering
	\caption{
		The ablation study results on the Twitter15, Twitter16, and Weibo datasets. 
	}
	\setlength{\tabcolsep}{3mm}{
		\begin{tabular}{p{6.5cm}|c|c|c}
			\toprule
			\multirow{2}[3]{*}{Models} & Twitter15 & Twitter16 & Weibo  \\
			& Accuracy & Accuracy & Accuracy \\
			\midrule
			SMAN$_{base}$    & 0.929  & 0.935  & 0.951    \\
			w/o Publisher Credibility (PC)           & 0.887  & 0.913  & 0.930     \\
			w/o User Credibility (UC)           & 0.905  & 0.880  & 0.938    \\
			w/o Publisher and User Credibility (PUC)         & 0.863  & 0.851  & 0.911    \\
			\bottomrule
		\end{tabular}
	}
	\label{ablation_results}
	\vspace{-1\baselineskip}
\end{table}

We first evaluate the impact brought by the publishers' credibility prediction subtask. We can observe that the performance drops a lot without the PC. The publishers' credibility prediction subtask can exploit publishing relations between publishers and corresponding news to transfer the influence of publishers' credibility to news credibility, thus facilitating the fake news detection. The ablation results also prove it is very important to explicitly encode the credibility of publishers. 

Then, we analyze the influence of the user credibility prediction subtask. We can observe that the absence of UC also causes significant performances to decline on all datasets. Intuitively speaking, if a piece of news is reposted by many low-reputation users, its credibility will indeed be greatly reduced. Same as PC task, the users' credibility also can be transferred to news credibility by diffusion graph, and thereby it can improve the detection performance. 

Finally, we also find that the performance is much lower than the complete model SMAN after removing both publisher and user credibility prediction subtasks, which further proves that both tasks provide complementary information to each other. Thus, it is essential to jointly optimize the fake news detection and credibility prediction tasks.

\subsection{Early Detection}
For fake news detection task, one of the most essential targets is to detect fake news as soon as possible to intervene in time~\cite{Zhao_2015}. To answer EQ3, we compared different methods of different time delays, and the performance is evaluated by the accuracy obtained when we incrementally add data up to the checkpoint given the targeted time delay. 

\begin{figure}[!htbp]
	\centering 
	\subfigure[Twitter15]{
		\label{fig:subfig:a1} 
		\includegraphics[scale=0.45]{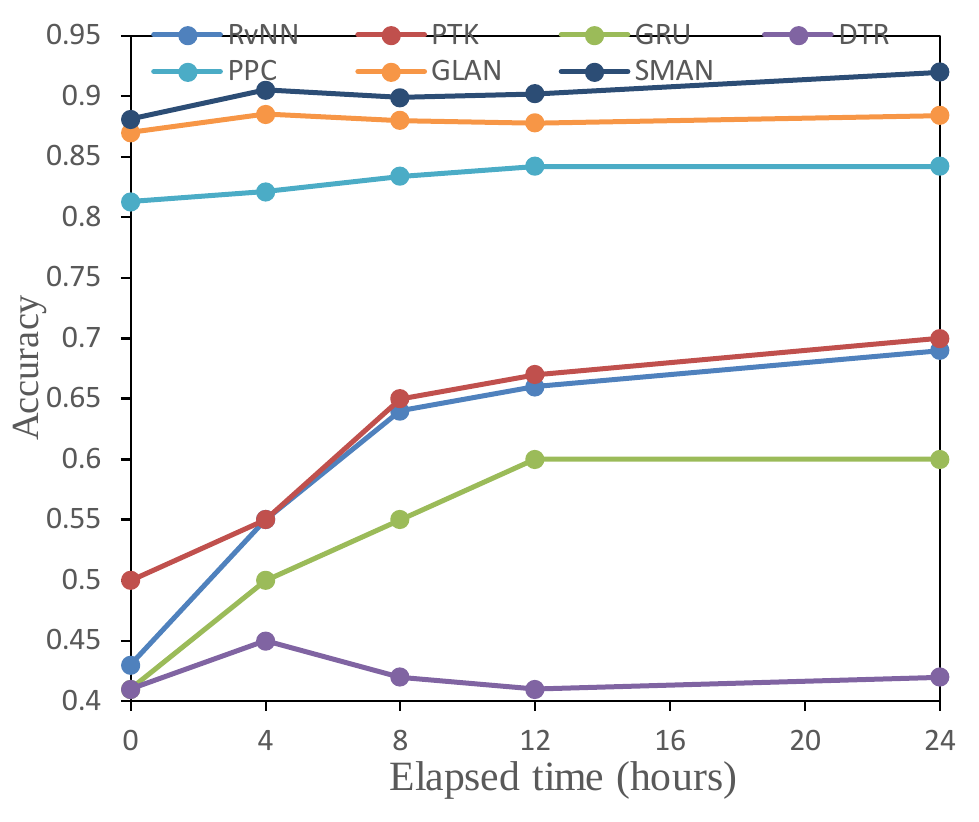} 
	}
	\subfigure[Twitter16]{
		\label{fig:subfig:b1} 
		\includegraphics[scale=0.45]{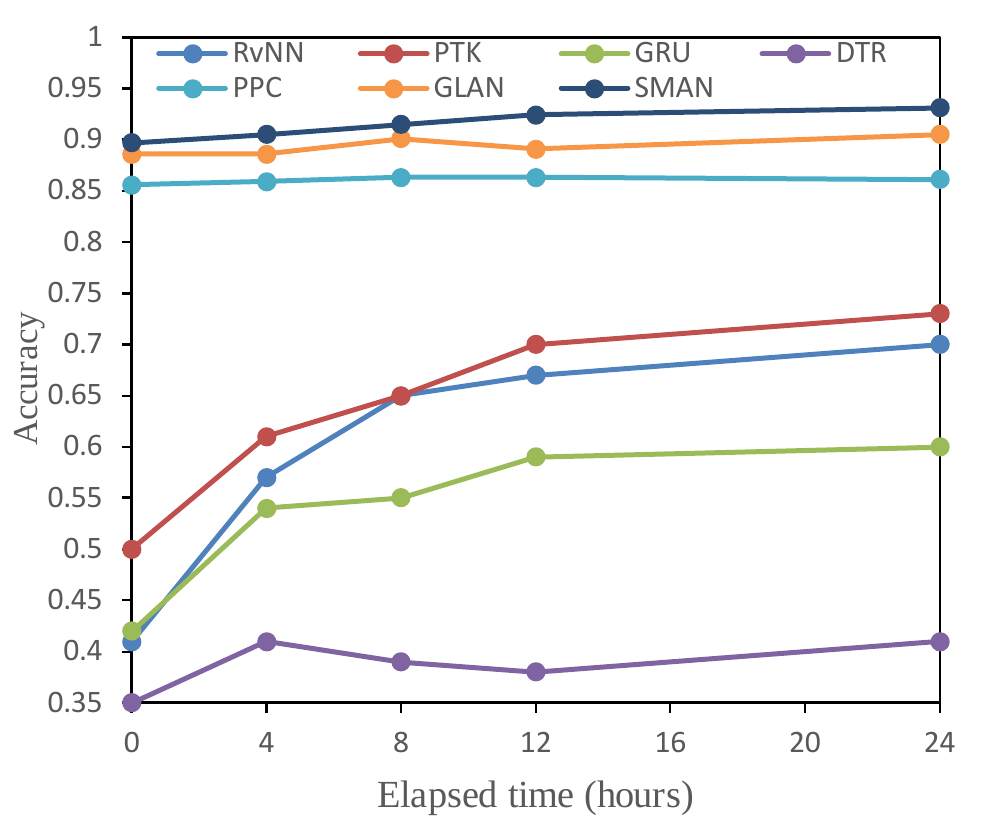} 
	}
	\subfigure[Weibo]{
		\label{fig:subfig:c1} 
		\includegraphics[scale=0.45]{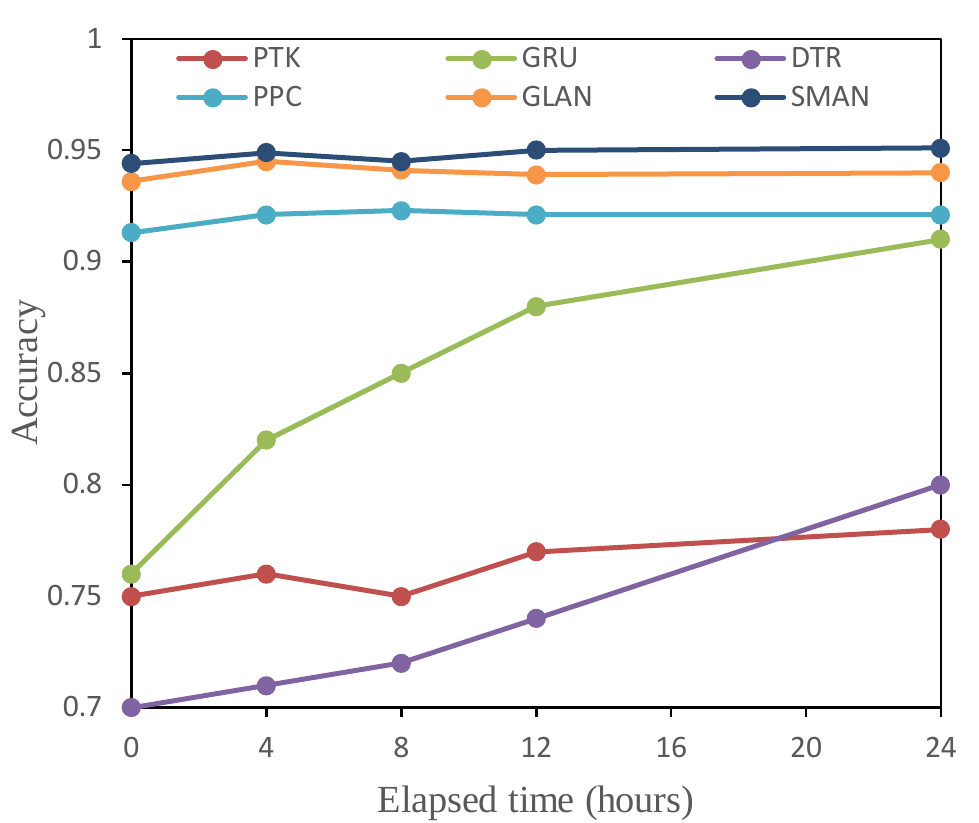} 
	}
	\caption{Results of early fake news detection on the Twitter15, Twitter16 and Weibo dataset.}
	\label{early_detection}
\end{figure}

By changing the time delays, the accuracy of several competitive models is shown in Figure~\ref{early_detection}. In 0 to 4 hours, SMAN significantly outperforms the tree-based methods and feature-based methods and achieves better performance over the state of the art method, indicating the superior early detection performance of SMAN. Particularly, SMAN achieves about 91\% accuracy on Twitter15 and Twitter16 datasets, and 95\% accuracy on Weibo within 4 hours, which is much faster than most of the baselines. 

After 8 hours, our model significantly surpasses the state of the art method. We can see that using more reposting relations will make the construction of the diffusion graph more complete and make the influence of credibility more easily transfer from publishers and users to news representations. Overall, the experimental results show that SMAN can not only improve the detection performance but also significantly reduce the time required for detection.

\section{Conclusion} \label{conclusion}  
This paper proposes a novel structure-aware multi-attention network, which combines news content, the heterogeneous graphs among publishers and users, and jointly optimizes the task of false news detection and user credibility prediction for early fake news detection. Different from most existing research extracting hand-crafted features or deep learning methods, we explicitly treat the credibility of publishers and users as a kind of weakly supervised information for facilitating fake news detection. Extensive experiments conducted on three real-world datasets show that the proposed model can significantly surpass other state-of-the-art models on both fake news classification and early detection task.

\section*{Acknowledgements}
We thank the anonymous reviewers for their feedback. This research is supported in part by the National Key Research and Development Program of China under Grant 2018YFC0806900.

\bibliographystyle{coling}
\bibliography{coling2020}

\end{document}